\documentclass[letterpaper,twocolumn,10pt]{article}
\usepackage{usenix,epsfig,endnotes,cite,comment}
\usepackage{algorithm}
\usepackage[noend]{algpseudocode}
\usepackage{amsmath}
\usepackage{amssymb}
\usepackage{multirow}
\usepackage{url}
\usepackage{flushend}
\usepackage{setspace}
\usepackage{xspace}
\usepackage{breakurl}
\usepackage{flushend}
\usepackage{subfig}
\usepackage{microtype}
\usepackage{makecell}
\usepackage[T1]{fontenc}

\newcommand{\ie}{i.e., }
\newcommand{\eg}{e.g., }

\newcommand{\scp}{symbolic interval analysis\xspace}
\newcommand{\iir}{iterative interval refinement\xspace}
\newcommand{\ia}{influence analysis\xspace}
\newcommand{\mono}{monotonicity\xspace}
\newcommand{\acasxu}{ACAS Xu\xspace}
\newcommand{\sys}{ReluVal\xspace}

\begin{document}
	
        \title{\Large \bf Formal Security Analysis of Neural Networks using Symbolic Intervals}
	\author{\rm Shiqi Wang, Kexin Pei, Justin Whitehouse, Junfeng Yang, and Suman Jana\\ 
Columbia University}
	
	\maketitle
	
	\subsection*{Abstract}
\label{sec:abst}
Due to the increasing deployment of Deep Neural Networks (DNNs) in real-world security-critical domains including autonomous vehicles and collision avoidance systems, formally checking security properties of DNNs, especially under different attacker capabilities, is becoming crucial. Most existing security testing techniques for DNNs try to find adversarial examples without providing any formal security guarantees about the non-existence of such adversarial examples. Recently, several projects have used different types of Satisfiability Modulo Theory (SMT) solvers to formally check security properties of DNNs. However, all of these approaches are limited by the high overhead caused by the solver.   

In this paper, we present a new direction for formally checking security properties of DNNs without using SMT solvers. Instead, we leverage interval arithmetic to compute rigorous bounds on the DNN outputs. Our approach, unlike existing solver-based approaches, is easily parallelizable. We further present symbolic interval analysis along with several other optimizations to minimize overestimations of output bounds. 

We design, implement, and evaluate our approach as part of \sys, a system for formally checking security properties of Relu-based DNNs. Our extensive empirical results show that \sys outperforms Reluplex, a state-of-the-art solver-based system, by 200 times on average. On a single 8-core machine without GPUs, within 4 hours, \sys is able to verify a security property that Reluplex deemed inconclusive due to timeout after running for more than 5 days. Our experiments demonstrate that symbolic interval analysis is a promising new direction towards rigorously analyzing different security properties of DNNs.



	\section{Introduction}
\label{sec:intro}

In the last five years, Deep Neural Networks (DNNs) have enjoyed tremendous progress, achieving or surpassing human-level performance in many tasks such as speech recognition~\cite{hinton2012deep}, image classifications~\cite{krizhevsky2012imagenet}, and game playing~\cite{silver2017mastering}.
We are already adopting DNNs in security- and mission-critical domains like collision avoidance and autonomous driving~\cite{bloom2017self, BaiduApollo}. 
For example, unmanned Aircraft Collision Avoidance System X (\acasxu), uses DNNs to predict best actions according to the location and the speed of the attacker/intruder planes in the vicinity.
It was successfully tested by NASA and FAA~\cite{NASAFAA, marston2015acas} and is on schedule to be installed in over 30,000 passengers and cargo aircraft worldwide~\cite{ACASXTech} and US Navy's fleets~\cite{MQ-4C}.

Unfortunately, despite our increasing reliance on DNNs, they remain susceptible to incorrect corner-case behaviors: \emph{adversarial examples}~\cite{szegedy2013intriguing}, with small, human-imperceptible perturbations of test inputs, unexpectedly and arbitrarily changing a DNN's predictions.
In a security-critical system like \acasxu, an incorrectly handled corner case can easily be exploited by an attacker to cause significant damage costing thousands of lives.

Existing methods to test DNNs against corner cases focus on finding
adversarial examples~\cite{pei2017deepxplore, goodfellow2014explaining, moosavi2016deepfool, carlini2017towards, papernot2016practical, kurakin2016adversarial, liu2016delving, nguyen2015deep, xu2016automatically} without providing formal guarantees about the non-existence of adversarial inputs even within very small input ranges. In this paper, we focus on the problem of formally checking that a DNN never violates a security property (\eg no collision) for any malicious input provided by an attacker within a given input range (\eg for attacker aircraft's speeds between $0$ and $500$ mph).  

Due to non-linear activation functions like ReLU, the general function computed by a DNN is highly non-linear and non-convex. Therefore it is difficult to estimate the output range accurately.
To tackle these challenges, all prior works on the formal security analysis of neural networks~\cite{katz2017reluplex, ehlers2017formal, huang2017safety, carlini2017ground} rely on different types of Satisfiability Modulo Theories (SMT) solvers and are thus severely limited by the efficiency of the solvers.

We present \sys, a new direction for formally checking security properties of DNNs without using SMT solvers. 
Our approach leverages interval arithmetic~\cite{intervalbook} to compute rigorous bounds on the outputs of a DNN. 
Given the ranges of operands (\eg $a_1 \in [0, 1]$ and $a_2 \in [2,3]$), interval arithmetic computes the output range efficiently using only the lower and upper bounds of the operands (\eg $a_2 - a_1 \in [1, 3]$ because $2-1=1$ and $3-0=3$).
Compared to SMT solvers, we found interval arithmetic to be significantly more efficient and flexible for formal analysis of a DNN's security properties.  

Operationally, given an input range $X$ and security property $P$, \sys propagates it layer by layer to calculate the output range, applying a variety of optimizations to improve accuracy.
\sys finishes with two possible outcomes: (1) a formal guarantee that no value in $X$ violates $P$ (``secure'');  and (2) an adversarial example 
in $X$ violating $P$ (``insecure''). 
Optionally, \sys can also guarantee that no value in a set of subintervals of $X$ violates $P$ (``secure subintervals'') and that all remaining subintervals each contains at least one concrete adversarial example of $P$ (``insecure subintervals''). 


A key challenge in \sys is the inherent overestimation caused by the input dependencies~\cite{de2004affine, intervalbook} when interval arithmetic is applied to complex functions.
Specifically, the operands of each hidden neuron depend on the same input to the DNN, but interval arithmetic assumes that they are independent and may thus compute an output range much larger than the true range. For example, consider a simplified neural network in which input $x$ is fed to two neurons that compute $2x$ and $-x$ respectively, and the intermediate outputs are summed to generate the final output $f(x)=2x-x$.  If the input range of $x$ is $[0,1]$, the true output range of $f(x)$ is $[0,1]$.
However, naive interval arithmetic will compute the range of $f(x)$ as $[0,2]-[0,1] = [-1,2]$, introducing a huge overestimation error. Much of our research effort focuses on mitigating this challenge; below we describe two effective optimizations to tighten the bounds.

First, \sys uses \emph{symbolic intervals} whenever possible to track the symbolic lower and upper bounds of each neuron. In the preceding example, \sys tracks the intermediate outputs symbolically ($[2x, 2x]$ and $[-x, -x]$ respectively) to compute the range of the final output as $[x, x]$.
When propagating symbolic bound constraints across a DNN, \sys correctly handles non-linear functions such as ReLUs and calculates proper symbolic upper and lower bounds.
It concretizes symbolic intervals when needed to preserve a sound approximation of the true ranges.  Symbolic intervals enable \sys to accurately handle input dependencies, reducing output bound estimation errors by 85.67\% compared to naive extension based on our evaluation.

Second, when the output range of the DNN is too large to be conclusive, \sys iteratively bisects the input range and repeats the range propagation on the smaller input ranges.  We term this optimization \emph{iterative interval refinement} because it is in spirit similar to abstraction
refinement~\cite{ball2002s,henzinger2002lazy}. 
Interval refinement is also amenable to massive parallelization, an additional advantage of \sys over hard-to-parallelize SMT solvers. 

Mathematically, we prove that interval refinement on DNNs always converges in finite steps as long as the DNN is Lipschitz continuous which is true for any DNN with finite number of layers. Moreover, lower values of Lipschitz constant result in faster convergence. Stable DNNs are known to have low Lipschitz constants~\cite{szegedy2013intriguing} and therefore the interval refinement algorithm can be expected to converge faster for such DNNs.
To make interval refinement even more efficient, \sys uses additional optimizations that analyze how each input variable influences the output of a DNN by computing each layer's gradients to input variables.
For instance, when bisecting an input range, \sys picks the input variable range that influences the output the most. Further, it looks for input variable ranges that influence the output monotonically, and uses only the lower and upper bounds of each such range for sound analysis of the output range, avoiding splitting any of these ranges.

We implemented \sys using around 3,000 line of \texttt{C} code.
We evaluated \sys on two different DNNs, \acasxu and an MNIST network, using $15$ security properties (out of which 10 are the same ones used in~\cite{katz2017reluplex}). Our results show that \sys can provide formal guarantees for all 15 properties, and is on average $200$ times faster than Reluplex, a state-of-the-art DNN verifier using a specialized solver~\cite{katz2017reluplex}. \sys is even able to prove a security property within 4 hours that Reluplex~\cite{katz2017reluplex} deemed inconclusive due to timeout after 5 days. For MNIST, \sys verified 39.4\% out of 5000 randomly selected test images to be robust against up to $|X|_{\infty}\leq 5$ attacks.

This paper makes three main contributions.

\begin{itemize}
\item To the best of our knowledge, \sys is the first system that leverages interval arithmetic to provide formal guarantees of DNN security.

\item Naive application of interval arithmetic to DNNs is ineffective.  We present two optimizations -- symbolic intervals and iterative refinement -- that significantly improve the accuracy of interval arithmetic on DNNs.

\item We designed, implemented, evaluated our techniques as part of \sys and demonstrated that it is on average 200$\times$ faster than Reluplex, a state-of-the-art DNN verifier using a specialized solver~\cite{katz2017reluplex}.
\end{itemize}

	\section{Background}
\label{sec:background}

\subsection{Preliminary of Deep Learning}
\label{subsec:prelim_dl}

A typical feedforward DNN can be thought of as a function $f: \mathbb{X}\rightarrow \mathbb{Y}$ mapping inputs $x\in \mathbb{X}$ (\eg images, texts) to outputs $y\in \mathbb{Y}$ (\eg labels for image classification, texts for machine translation).
Specifically, $f$ is composed of a sequence of parametric functions $f(x;w)=f_l(f_{l-1}(\cdots f_2(f_1(x;w_1);w_2)\cdots w_{l-1}),w_l)$, where $l$ denotes the number of layers in a DNN, $f_k$ denotes the corresponding transformation performed by $k$-th layer, and $w_k$ denotes the weight parameters of $k$-th layer.
Each $f_{k\in{1,...l}}$ performs two operations: (1) a linear transformation of its input (\ie either $x$ or the output from $f_{k-1}$) denoted by $w_k \cdot f_{k-1}(x)$, where $f_0(x)=x$ and $f_{k\neq0}(x)$ is the output of $f_k$ denoting intermediate output of layer $k$ while processing $x$, 
and (2) a nonlinear transformation $\sigma(w_k \cdot f_{k-1}(x))$ where $\sigma$ is the nonlinear activation function.
Common activation functions include sigmoid, hyperbolic tangent, or ReLU (Rectified Linear Unit)~\cite{nair2010rectified}.
In this paper, we focus on DNNs using ReLU ($Relu(x) = max(0,x)$) as the activation function as it is one of the most popular ones used in the modern state-of-the-art DNN architectures~\cite{he2016deep, huang2017densely, szegedy2016rethinking}.

\subsection{Threat Model}
\label{subsec:threat_model}

\noindent\textbf{Target system.}
In this paper, we consider all types of security-critical systems, \eg airborne collision avoidance system for unmanned aircraft like ACAS Xu~\cite{marston2015acas}, which use DNNs for decision making in the presence of an adversary/intruder. DNNs are becoming increasingly popular in such systems due to better accuracy and less performance overhead than traditional rule-based systems~\cite{julian2016policy}. For example, an aircraft collision avoidance system's decision-making process can use DNNs to predict the best action based on sensor data of the current speed and course of the aircraft, those of the adversary, and distances between the aircraft and nearby intruders. 

\begin{figure}[!hbt]
	\centering
	\includegraphics[width=.9\columnwidth]{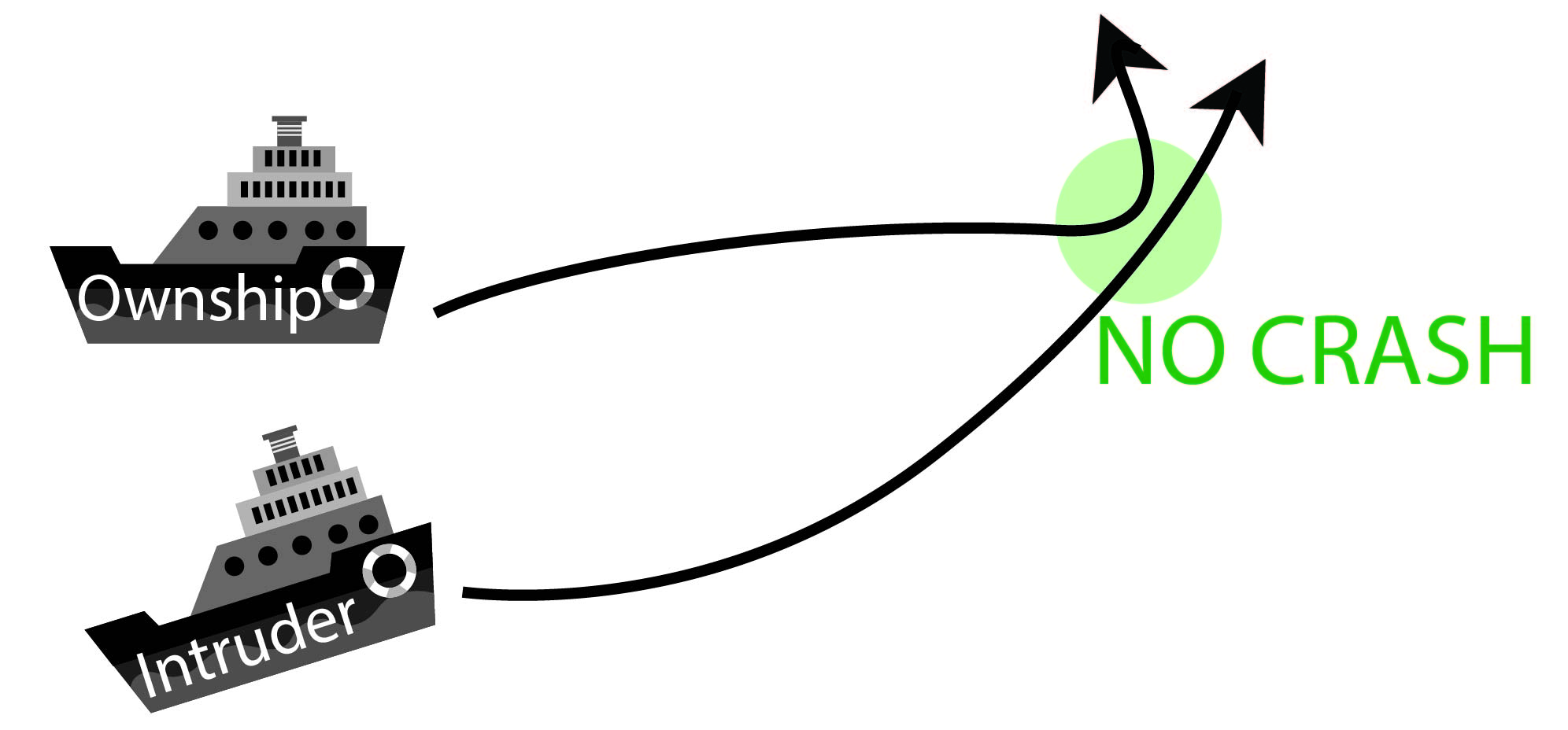}
	\includegraphics[width=.9\columnwidth]{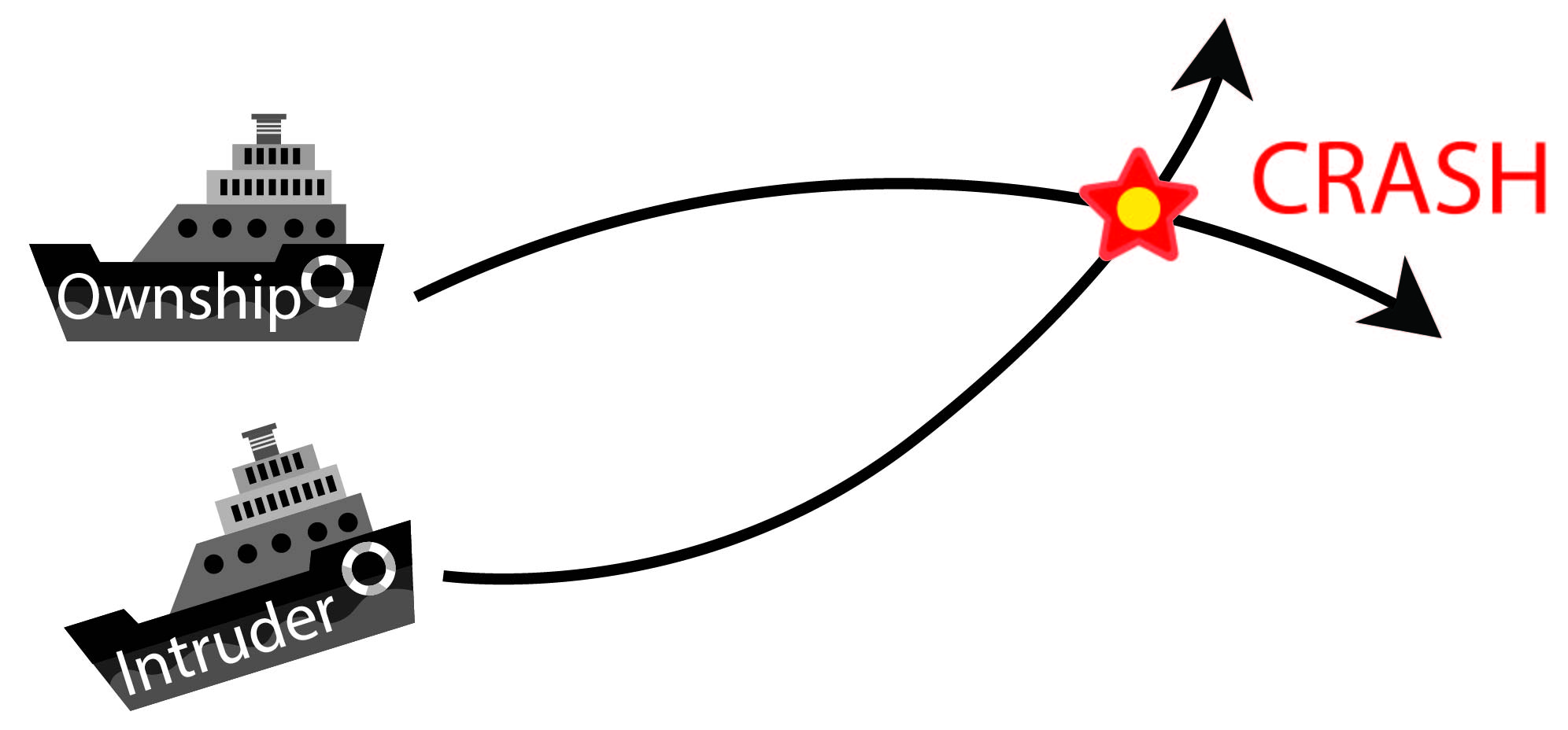}
	\caption{The DNN in the victim aircraft (ownship) should predict a left turn (upper figure) but unexpectedly advises to turn right and collides with the intruder (lower figure) due to the presence of adversarial inputs (e.g., if the attacker approaches at certain angles).}
	\label{fig:threat_model}
\end{figure}

\noindent\textbf{Security properties.}
In this paper, we focus on input-output-based security properties of DNN-based systems that ensure the correct actions in the presence of adversarial inputs within a given range. Input-output properties are well suited for the DNN-based systems as their decision logic is often opaque even to their designers. Therefore, unlike traditional programs, writing complete specifications involving internal states is often hard.

For example, consider a security property that tries to ensure that a DNN-based car crash avoidance system predicts the correct steering angle in the presence of an approaching attacker vehicle: it should steer left if the attacker approaches it from right. In this setting, even though the final decision is easy to predict for humans, the correct outputs for the internal neurons are hard to predict even for the designer of the DNN.

\noindent\textbf{Attacker model.}
We assume that the inputs an adversary can provide are bounded within an interval specified by a security property. For example, an attacker aircraft has a maximum speed (e.g., it can only move between 0 and 500 mph).  Therefore, the attacker is free to choose any value within that range. This attacker model is, in essence, similar to the ones used for adversarial attacks on vision-based DNNs where the attacker aims to search for visually imperceptible perturbations (within certain bound) that, when applied on the original image, makes the DNN predict incorrectly. Note that, in this setting, the imperceptibility is measured using a $L_p$ norm. 
Formally, given a computer vision DNN $f$, the attacker solves following optimization problem: $min(L_p(x'-x))$ such that $f(x)\neq f(x')$, where $L_p(\cdot)$ denotes the $p$-norm and $x'-x$ is the perturbation applied to original input $x$. In other words, the security property of a vision DNN being robust against adversarial perturbations  can be defined as: for any $x'$ within a $L$-distance ball of $x$ in the input space, $f(x)=f(x')$.

Unlike the adversarial images, we extend the attacker model to allow different amounts of perturbations to different features. Specifically, instead of requiring overall perturbations on input features to be bounded by L-norm, our security properties allow different input features to be transformed within different intervals. Moreover, for DNNs where the outputs are not explicit labels, unlike adversarial images, we do not require the predicted label to remain the same. We support properties specifying arbitrary output intervals.

\noindent\textbf{An example.}
As shown in Figure~\ref{fig:threat_model}, normally, when the distance (one feature of the DNN) between the victim ship (ownship) and the intruder is large, the victim ship advisory system will advise left to avoid the collision and then advise right to get back to the original track. 
However, if the DNN is not verified, there may exist one specific situation where the advisory system, for certain approaching angles of the attacker ship, advises the ship incorrectly to take a right turn instead of left, leading to a fatal collision. If an attacker knows about the presence of such an adversarial case, he can specifically approach the ship at the adversarial angle to cause a collision.

\subsection{Interval Analysis}
\label{subsec:interval_analysis}
Interval arithmetic studies the arithmetic operations on intervals rather than concrete values.
As discussed above, since (1) the DNN safety property checking requires setting input features within certain ranges and checking the output ranges for violations, and (2) the DNN computations only include additions and multiplications (linear transformations) and simple nonlinear operations (\eg ReLUs), interval analysis is a natural fit to our problem. We provide some formal definitions of interval extensions of functions and their properties below. We use these definitions in Section~\ref{sec:prove} for demonstrating the correctness of our algorithm. 

Formally, let $x$ denote a concrete real value and $X := [\underline{X}, \overline{X}]$ denote an interval, where $\underline{X}$ is the lower bound, and $\overline{X}$ is the upper bound. An \textit{interval extension} of a function $f(x)$ is a function of intervals $F$ such that, for any $x \in X$, $F([x, x]) = f(x)$. The ideal interval extension $F(X)$ approaches the image of $f$, $f(X):=\{f(x):x \in X\}$.

Let $f(X_1, X_2, ... , X_d):=　\{f(x_1, x_2,...,x_d): x_1 \in X_1, x_2 \in X_2,...,x_d \in X_d \}$ where d is the number of input dimensions. An interval valued function $F(X_1,X_2,...,X_d)$ is \textit{inclusion isotonic} if, when $Y_i\subseteq X_i \text{ for } i=1,...,d$, we have $$F(Y_1, Y_2,...,Y_d)\subseteq F(X_1,X_2,...,X_d)$$

An interval extension function $F(X)$ that is defined on an interval $X_0$ is said to be \textit{Lipschitz continuous} if there is some number $L$ such that:$$ \forall X \subseteq X_0, w(F(X))\leq L \cdot w(X)$$
where $w(X)$ is the width of interval $X$, and $X$ here denotes $X = (X_1, X_2,..., X_d)$, a vector of intervals~\cite{intervalbook}.
	
	\section{Overview}
\label{sec:overview}

\begin{figure}[!htb]
	\centering
	\includegraphics[width=.9\columnwidth]{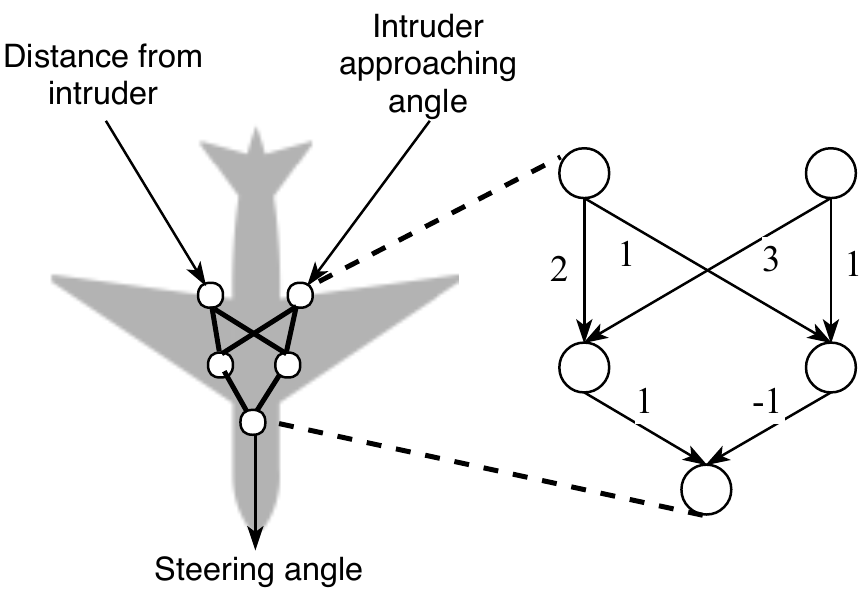}
	\caption{Running example to demonstrate our techniques.}
	\label{fig:main}
\end{figure}

Interval analysis is a natural fit to the goal of verifying safety properties in neural networks as we have discussed in Section~\ref{subsec:interval_analysis}.
Naively, by setting input features as intervals, we could follow the same arithmetic performed in the DNN to compute the output intervals.
Based on the output intervals, we can verify if the input perturbations will finally lead to violations or not (\eg output intervals go beyond a certain bound). Note that, lack of violations indicates the safety property is verified to be safe due to over-approximations.

However, naively computing output intervals in this way suffers from high errors as it computes extremely loose bounds due to the \textit{dependency problem}.
In particular, it can only get a highly conservative estimation of the output range, which is too wide to be useful for checking any safety property.
In this section, we first demonstrate the dependency problem with a motivating example using naive interval analysis. Next, based on the same example, we describe how the techniques described in this paper can mitigate this problem.

\begin{figure*}[!htb]
	\centering
	\includegraphics[width=\linewidth]{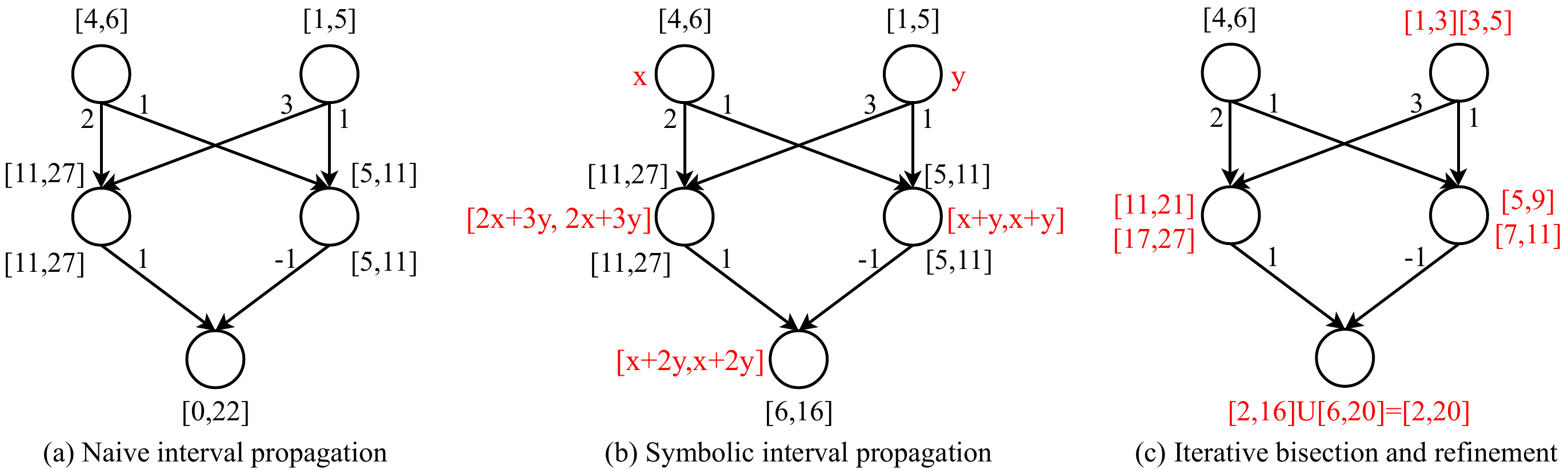}
	\caption{Examples showing (a) naive interval extension where the output interval is very loose as it ignores the inter-dependency of the input variables, (b) using \scp to keep track of some of the dependencies, and (c) using bisection to reduce the over-approximation error.}
	\label{fig:interval_optimize}
\end{figure*}

\noindent\textbf{A working example.} 
We use a small motivating example shown in Figure~\ref{fig:main} to illustrate the inter-dependency problem and our techniques in dealing with this problem in Figure~\ref{fig:interval_optimize}.

Let us assume that the sample NN is deployed in an unmanned aerial vehicle taking two inputs (1) distance from the intruder and (2) intruder approaching angle, while producing the steering angle as output. The NN has five neurons arranged in three layers. The weights attached to each edge is also shown in Figure~\ref{fig:interval_optimize} .

Assume that we aim to verify if the predicted steering angle is safe by checking a property that the steering angle should be less than 20 if the distance from the intruder is in $[4,6]$ and the possible angle of approaching intruder is in $[1,5]$.

Let $x$ denote the distance from an intruder and $y$ denote the approaching angle of the intruder.
Essentially, given $x\in[4,6]$ and $y\in[1,5]$, we aim to assert that $f(x,y)\in[-\infty,20]$.
Figure~\ref{fig:interval_optimize}a illustrates the naive interval propagation in this NN. 
By performing the interval multiplications and additions, along with applying the ReLU activation functions, we get the output interval to be $[0,22]$.
Note that this is an overestimation because the upper bound 22 cannot be achieved: it can only appear when the left hidden neuron outputs 27 and the right one outputs 5.
However, for the left hidden neuron to output 27, the conditions $x=6$ and $y=5$ have to be satisfied. 
Similarly, for the right hidden neuron to output 5, the conditions $x=4$ and $y=1$ have to be satisfied.
These two conditions are contradictory and therefore cannot be satisfied simultaneously and therefore the final output 22 can never appear.
This effect is known as the \textit{dependency problem}~\cite{intervalbook}.

As we have defined that a safe steering angle must be less than or equal to 20, we cannot guarantee non-existence of violations, as the steering angle can have a value as high as 22 according to the naive interval propagation described above.

\noindent\textbf{Symbolic interval propagation.}
Figure~\ref{fig:interval_optimize}b demonstrates how we maintain the \textit{symbolic intervals} to preserve as much dependency information as we can while propagating the bounds through the NN layers. In this paper, we only keep track of linear symbolic bounds and concretize the bounds when it is not possible to maintain accurate linear bounds. We compute the final output intervals using the corresponding symbolic equations. Our approach helps in significantly cutting down the over-approximation errors.

For example, in the current example, the intermediate neurons update their symbolic lower and upper bounds to be $2x+3y$ and $x+y$, denoting the operations performed by the previous linear transformations (taking the dot product of the input and weight parameters).
As we also know $2x+3y>0$ and $x+y>0$ for the given input range $x\in[4,6]$ and $y\in[1,5]$, we can safely propagate the symbolic intervals through the ReLU activation functions.


In the final layer, the propagated bound will be $[x+2y, x+2y]$, where we can finally compute the concrete interval $[6,16]$. This is tighter than the naive baseline interval $[0,22]$ and can be used to verify the property that the steering angle will be less than 20.

In summary, \textit{symbolic interval propagation} explicitly represents the intermediate computations of each neuron in terms of the symbolic intervals that encode the inter-dependency of the inputs to minimize overestimation.  

However, in more complex cases, there might be intermediate neurons with symbolic bounds whose possible values can potentially be negative. For such cases, we can no longer keep the symbolic interval using a linear equation while passing it through a ReLU. Therefore, we concretize their upper and lower bounds and ignore their dependencies. To minimize the errors caused by such cases, we introduce another optimization, \textit{iterative refinement}, as described below. As shown in Section~\ref{sec:eval}, we can achieve very tight bounds by combining these two techniques.

\noindent\textbf{Iterative refinement.}
Figure~\ref{fig:interval_optimize}c illustrates another optimization that we introduce for mitigating the dependency problem. Here, we leverage the fact that the dependency error for Lipschitz continuous functions decreases as the width of intervals decreases (any DNN with a finite number of layers is Lipschitz continuous as shown in Section~\ref{lip_proof}).
Therefore, we can bisect the input interval by evenly dividing the interval into the union of two consecutive sub-intervals and reduce the overestimation.
The output bound can thus be tightened as shown in the example. The interval becomes $[2,20]$, which proves the non-existence of the violation.
Note that we can iteratively refine the output interval by repeated splitting of the input intervals. Such operations are highly parallelizable as the split sub-intervals can be checked independently (Section~\ref{sec:eval}). In Section~\ref{sec:prove}, we provide a proof that the iterative refinement can effectively reduces the width of the output range to an arbitrary precision within finite steps for any Lipschitz continuous DNN.

    \vspace{-.4cm}
\section{Proof of Correctness}
\label{sec:prove}

Section~\ref{sec:overview} demonstrates the basic idea of naive interval extension and the optimization of iterative refinement. 
In this section, we give the detailed proof about the correctness of interval analysis/estimation on DNNs, also known as interval extension estimation, and the convergence of iterative refinement.
The proofs are based on two aforementioned properties of neural networks: \textit{inclusion isotonicity} and \textit{Lipschitz continuity}.
In general, the correctness guarantee of interval extension holds for most finite DNNs while the convergence guarantee requires Lipschitz continuity.
In the following, we give the proof of correctness for two most important techniques we use throughout the paper, but the proof is generic and works for our other optimizations such as \scp, influence analysis and monotonicity as described in Section~\ref{sec:methodology}. 

Let $f$ denote an NN and $F$ denote its naive interval extension. We define the naive interval extension as a function $F(X)$ that (1) satisfies for all $x \in X, F([x,x]) = f(x)$ and (2) that only involves naive interval operations during interval variable representations. For all the other types of interval extensions, they can be easily analyzed based on the following proof.

\subsection{Correctness of Overestimation}

We are going to demonstrate that, for the naive interval extension of $f$, $F$ always overestimates the theoretically tightest output range $f$. 
According to our definition of inclusion isotonicity described in Section~\ref{sec:background}, it suffices to prove that the naive interval extension of an NN is inclusion isotonic. Note that we only consider neural networks with ReLUs as activation functions for the following proof, but the proof can be easily extended to other popular activation functions like tanh or sigmoid. 

First, we need to demonstrate that $F$ is inclusion isotonic.
Because ReLU is monotonic, so we can simply consider its interval extension to be $Relu_I(X) := [max(0,\underline{X}), max(0, \overline{X})]$. 
Therefore, $\forall Y\subset X$, we have $max(0,\underline{X})\leq max(0,\underline{Y})$ and $max(0,\overline{X})\geq max(0,\overline{Y})$ so that its interval extension $Relu(Y)\subseteq Relu(X)$. 
Most common activation functions are inclusion isotonic. We refer interested readers to~\cite{intervalbook} for a list of common functions that are inclusion isotonic. 

We note that $f(X)$ is a composition of activation functions and linear functions. And we also see that linear functions, as well as common activation functions, are inclusion isotonic\cite{intervalbook}. Because any combinations of inclusion isotonic functions are still inclusion isotonic, thus, we have that the interval representation $F(X)$ of $f(X)$ is inclusion isotonic.

Next, we show for arbitrary $X = (X_1, \dots, X_d)$, that: $$f(X) \subseteq F(X)$$
Applying the previously shown inclusion isotonicity properties of $F(X)$, we get:

\[
f(X_1, \dots, X_d) = \bigcup_{(x_1, \dots, x_d) \in X}\{f(x_1, \dots, x_d)\}
\]
\[
= \bigcup_{(x_1, \dots, x_d) \in X}F([x_1, x_1], \dots, [x_d,x_d])
\]
Now, for any such $(x_1, \dots, x_d) \in X$, we have $F([x_1, x_1], \dots, [x_d,x_d]) \subseteq F(X_1, \dots, X_d)$, since $([x_1, x_1], \dots, [x_d,x_d]) \subseteq (X_1, \dots, X_d)$, and $F(X)$ is inclusion isotonic. We thus get:
\begin{equation}
    \bigcup_{(x_1, \dots, x_d) \in X}F([x_1, x_1], \dots, [x_d,x_d]) \subseteq F(X_1, \dots, X_d)
    \label{eq:ii}
\end{equation}
which is exactly the desired result.

Now, we get the result shown in Equation~\ref{eq:ii} that for all input $X$, the interval extension of $f$, $F(X)$, always contains the true codomain (theoretically tightest bound) for $f(X)$.

\subsection{Convergence in Finite Number of Splits}
\label{lip_proof}
Now we see that the naive interval extension of $f$ is an overestimation of true output. Next, we show that iteratively splitting input is an effective way to refine and reduce such overestimated error. Empirically, we can see finite number of splits allow us to approximate $f$ with $F$ with arbitrary accuracy, this is guaranteed by Lipschitz continuity property of NNs.

First, we need to prove $F$ is Lipschitz continuous.
It is straightforward to show that many common activation functions are Lipschitz continuous~\cite{intervalbook}. Here, we show the natural interval extension $\mathrm{Relu}_I$ is Lipschitz continuous, with a Lipschitz constant $L := 1$. We see, for any input interval $X$:
\[
w(\mathrm{Relu}_I(X)) = max(\overline{X}, 0) - max(\underline{X}, 0) \]
\[
\leq max(\overline{X}, 0) - \underline{X} \leq \overline{X} - \underline{X} = w(X)
\]
Thus, the interval extension $\mathrm{Relu}_I$ of ReLU is Lipschitz continuous.
As the NN is a finite composition of Lipschitz continuous functions, its interval extension $F$ is still Lipschitz continuous as well~\cite{intervalbook}.

Now we demonstrate that by splitting input $X$ into $N$ smaller pieces and taking the union of their corresponding outputs, we can achieve a refined output estimation with at least $N$ times smaller overestimation error. We define an $N$-split uniform subdivision of input $X =(X_1,...,X_d)$ as a collection of sets $X_{i,j}$:$$X_{i,j}:=[X_i+(j-1)\frac{w(X_i)}{N}, X_i+j\frac{w(X_i)}{N}]$$ where $i \in 1, \dots, d$ and $j \in 1, \dots N$. We note that this is exactly a partition of each $X_i$ into $N$ pieces of equivalent width such that $\forall i,j $, $w(X_{i,j}) = w(X_i)/N$ and $X_i = \bigcup_{j = 1}^NX_{i,j}$.
We then define a refinement of $F$ over $X$ with $N$ splits as: $$F^{(N)}(X) := \bigcup^N_{i=1}F(X_{1,i}, \dots, X_{d, i})$$

Finally, we define the range of overestimated error created by naive interval extension on an NN after $N$-split refinement as $w(E^{(N)}(X))$: $$w(E^{(N)}(X)) := w(F^{(N)}(X))-w(f(X))$$

Because $F$ is Lipschitz continuous, Theorem 6.1 in ~\cite{intervalbook} gives us the following result:

\begin{equation}
    w(E^{(N)}(X)) \leq 2L \cdot w(X)/N
    \label{eq:lip}
\end{equation}

Equation~\ref{eq:lip} shows the error width of the $N$-split refinement $w(E^{(N)}(X))$ converges to 0 linearly as we increase $N$. That is, we can achieve arbitrary accuracy when using $N$-split refinement to approximate $f(X)$ with sufficiently large $N$.

	\section{Methodology}
\label{sec:methodology}



\begin{figure}[!htb]
    \centering
    \includegraphics[width=.95\columnwidth]{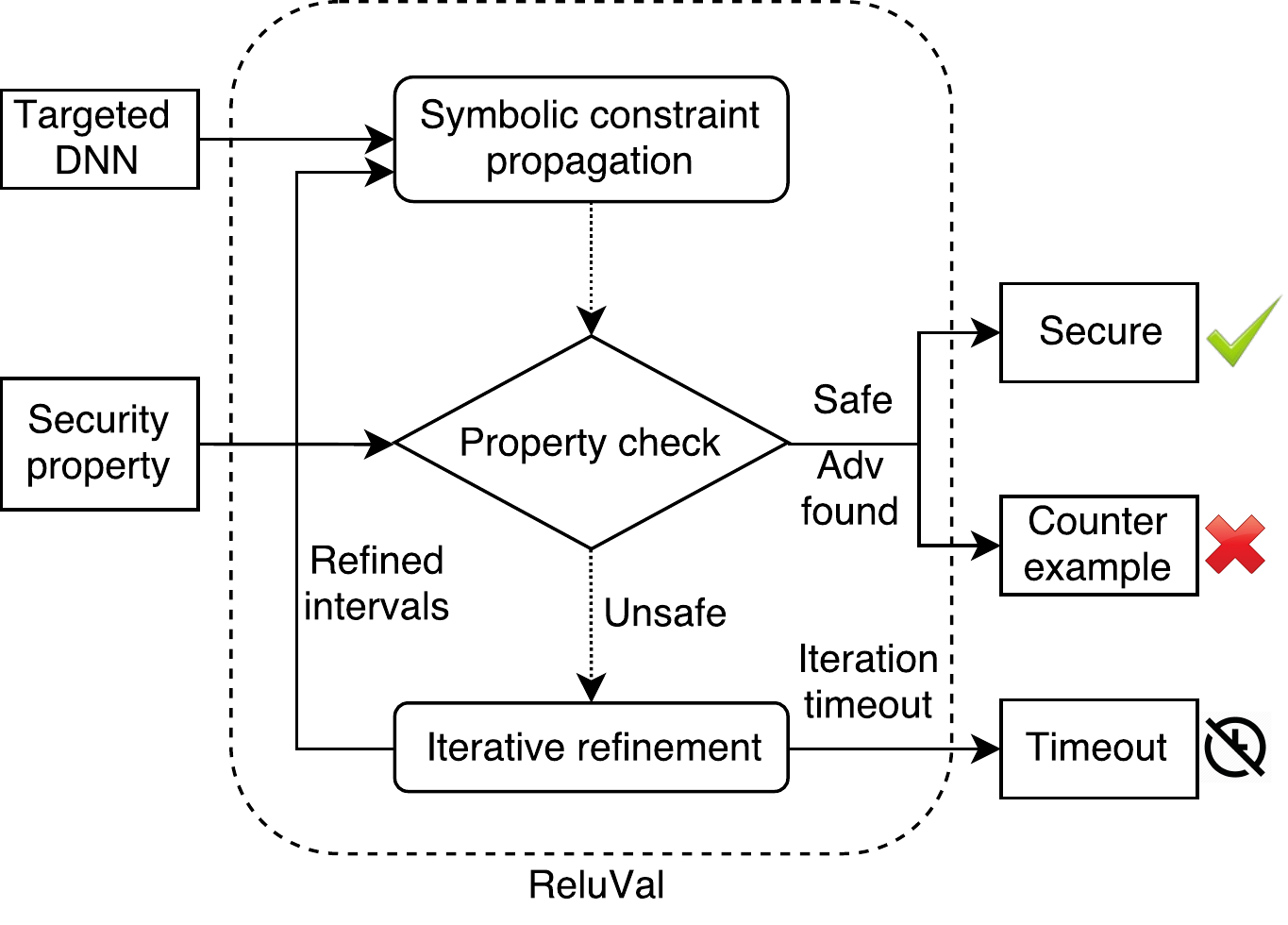}
    \caption{Workflow of \sys in checking security property of DNN.}
    \label{fig:workflow}
\end{figure}

Figure~\ref{fig:workflow} shows the main workflow along with the different components of \sys. 
Specifically, \sys uses \scp to get a tight estimation of the output ranges based on the input ranges. 
It declares a security property as verified if the estimated output interval is tight enough to satisfy the property.
If the output interval shows potential existence of violations, \sys randomly samples a few points from the interval and check for violations. If any adversarial case is detected, \ie a concrete input violating the security property, it outputs this as a counterexample. 
Otherwise, \sys uses \iir to further tighten the output interval to approach the theoretically tightest bound and repeats the same process described above. 
Once the number of iterations reaches a preset threshold, \sys outputs timeout denoting it cannot verify the security property.

As discussed in Section~\ref{sec:overview}, simple interval extension only obtains loose/conservative intervals due to input dependency problem. Below, we describe the details of the optimizations we propose to further tighten the bounds.

\subsection{Symbolic Interval Propagation}

Symbolic Interval propagation is one of our core contributions to mitigate the input dependency problem and tighten the output interval estimation.
If a DNN would only consist of linear transformations, keeping symbolic equation throughout the intermediate computations of a DNN can perfectly eliminate the input dependency errors. 

However, as shown in Section~\ref{sec:overview}, passing an equation through a ReLU node essentially involves dropping the equation and replacing it with 0 if the equation can evaluate to a negative value for the given input range. Therefore, we keep the lower and upper bound equations $(Eq_{up}, Eq_{low})$ for as many neurons as we can and only concretize as needed. 

\begin{algorithm}[!hbt]
	\caption{Symbolic interval analysis}
    \label{alg:forward}
	\scalebox{0.75}{
		\begin{tabular}{|llp{9em}|}
			\hline
			\textbf{Inputs}: & \textbf{network} $\leftarrow$ tested neural network & \  \\ 
			\ & \textbf{input} $\leftarrow$ input interval& \ \\
			\hline
		\end{tabular}}
	\small
	\begin{spacing}{0.9}
	\begin{algorithmic}[1]
		\State Initialize $eq = (eq_{up}, eq_{low})$;
		\State // cache mask matrix needed in backward propagation
		\State $R$[numLayer][layerSize];
		\State // loops for each layer
		\For {layer = 1 $\mathbf{to}$ numlayer}
			\State // matmal equations with weights as interval;
			\State $eq$= weight $\bigotimes$ $eq$;
			\State // update the output ranges for each node
			\If {layer != lastLayer}
				\For {i = 1 $\mathbf{to}$ layerSize[layer]}
					\If{$ \overline{eq_{up}[i]} \leq 0$} 
						\State // Update to 0
						\State R[layer][i]=[0,0];\Comment{$\frac{d(relu(x))}{dx}=[0,0]$}
						\State $eq_{up}[i]$ =  $eq_{low}[i] = 0$;
					\ElsIf{$\underline{eq_{low}[i]}\geq$0}
						\State // Keep dependency
						\State R[layer][i]=[1,1];\Comment{$\frac{d(relu(x))}{dx}=[1,1]$}
					\Else 
						\State // Concretization
						\State R[layer][i]=[0,1]; \Comment{$\frac{d(relu(x))}{dx}=[0,1]$}
						\State $eq_{low}[i]=0$			
						\If{$\underline{eq_{up}[i]}\leq0$} 
							\State $eq_{up}[i] = \overline{eq_{up}[i]}$;	
						\EndIf
					\EndIf
				\EndFor
			\Else
				\State output = \{lower, upper\};
			\EndIf
		\EndFor
		\State \Return R, output;
		
	\end{algorithmic}
	\end{spacing}
\end{algorithm}

Algorithm~\ref{alg:forward} elaborates the procedure of propagating symbolic intervals/equations during the interval computation of a DNN.
We describe the core components and the details of this technique below.

\noindent\textbf{Constructing symbolic intervals.} Given a particular neuron $A$, (1) If $A$ is in the first layer, we can compute the symbolic bounds as: $$Eq_{up}^A(X)=Eq_{low}^A(X) = w_1 x_1 + ... + w_d x_d$$ where $x_1,...,x_d$ are the inputs and $w_1,...,w_d$ are the weights of the corresponding edges. 
(2) If $A$ belongs to the intermediate layer, we initialize the symbolic intervals of $A$'s output as: $$Eq_{up}^{A}(X) = W_{+}Eq_{up}^{A_{prev}}(X)+W_{-}Eq_{low}^{A_{prev}}(X)$$
$$Eq_{low}^A(X) = W_{+}Eq_{low}^{A_{prev}}(X)+W_{-}Eq_{up}^{A_{prev}}(X)$$
where $Eq_{up}^{A_{prev}}$ and $Eq_{low}^{A_{prev}}$ are the equations from last layer. $W_{+}$ and $W_{-}$ denote the positive and negative weights of current layer respectively. The output will be $[w_+a,w_+b]$ for multiplying positive weight parameters $w_{+}$ with an interval $[a,b]$. For the negative weight parameters, the output will be flipped in terms of $a$ and $b$, \ie  $[w_-b,w_-a]$.

\noindent\textbf{Concretization.} While passing a symbolic equation through the ReLU nodes, we evaluate the concrete value of the equation's upper and lower bounds $\overline{Eq_{up}(X)}$ and $\underline{Eq_{low}(X)}$. If $\underline{Eq_{low}(X)}>0$, then we pass the lower equation on to the next layer. Otherwise, we concretize it to be $0$. Similarly, if $\underline{Eq_{up}(X)}>0$, we pass the upper equation on to the next layer. Otherwise, we concretize it as $\overline{Eq_{up}(X)}$. 

\noindent\textbf{Correctness.} We first clarify three different output intervals: (1) theoretically tightest bound $f(X)$, (2) naive interval extension bound $F(X)$, and (3) symbolic bound $[\underline{Eq_{low}(X)}, \overline{Eq_{up}(X)}]$. 
We prove that the symbolic bound is a superset of theoretically tightest bound and a subset of naive interval extension bound:

\begin{equation}
    f(X)\subseteq [\underline{Eq_{low}(X)}, \overline{Eq_{up}(X)}] \subseteq F(X)
    \label{eq:scp_correct}
\end{equation}

For a given input range propagated to the output layer, it will involve both computing linear transformations and applying ReLUs. Symbolic interval analysis keeps the accurate bounds for linear transformations and uses concretization to handle non-linearity. Compared to theoretically tightest bound, the only approximation introduced during the symbolic propagation process is due to concretization while handling ReLU nodes, which is an over-approximation as shown before. Naive interval extension, on the other hand, is a degenerate version of \scp where it does not keep any symbolic constraints. Therefore, \scp over-approximates the theoretically tightest bound and, in turn, is over-approximated by naive interval extension as shown in Equation~\ref{eq:scp_correct}.


\subsection{Iterative Interval Refinement}
\label{subsec:iir}

While \scp helps in computing relatively tight bounds, the estimated output intervals for complex networks may still not be tight enough for verifying properties, especially when the input intervals are comparably large and thus resulting in many concretizations. As discussed above in Section~\ref{sec:methodology}, for such cases, we resort to another technique, \iir.
In addition, we also propose two other optimizations, \ia and \mono, which further refine the estimated output ranges based on \iir.

\noindent\textbf{Baseline iterative refinement.} 
In Section~\ref{sec:prove}, we have proved that theoretically tightest bound could be approached by repeatedly splitting the input intervals. 
Therefore, we perform iterative bisections on each input interval $X_1,..., X_n$ until the output interval is tight enough to meet the security property, or time out, as shown in Figure~\ref{fig:workflow}.

\begin{figure}[!htb]
\includegraphics[width=0.9\columnwidth]{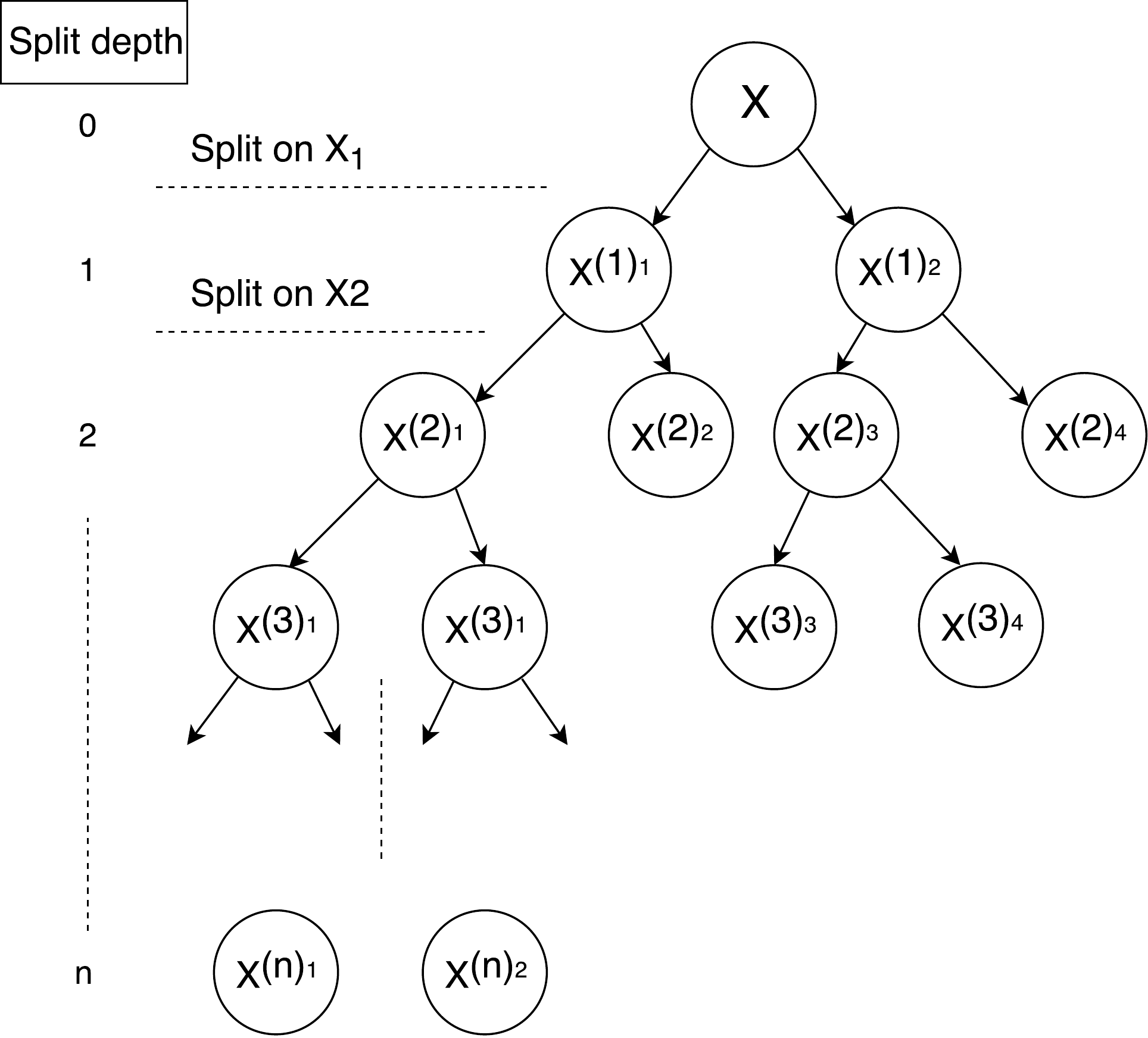}
\caption{A bisection tree with split depth of $n$. Each node represents a bisected sub-interval.}
\label{fig:bisectionTree}
\end{figure}

The iterative bisection process can be represented as a bisection tree as shown in Figure~\ref{fig:bisectionTree}. Each bisection on one input yields two children denoting two consecutive sub-intervals, the union of which computes the output bound for their parent. Here, $X^{(i)_j}$ means the $jth$ input interval with split depth $i$. 
After one bisection on $X^{(i)_j}$, it creates two children: $X^{(i+1)_{2j-1}} = \{X_1,...,[\underline{X_i}, \frac{\underline{X_i}+\overline{X_i}}{2}], ...,X_d\}$ and $X^{(i+1)_{2j}} = \{X_1,...,[\frac{\underline{X_i}+\overline{X_i}}{2}, \overline{X_i}],..., X_d\}$. 

To identify the existence of any adversarial example in the bisected input ranges, we sample a few input points (the current default is the middle point of each range) and verify if the concrete output leads to any property violation. If so, we output the adversarial example, mark this sub-interval as definitely containing adversarial examples, and conclude the analysis for this specific sub-interval. Otherwise, we repeat the \scp process for the sub-intervals. This default configuration is tailored towards deriving a conclusive answer of ``secure'' or ``insecure'' for the entire input intervals. Users of \sys can configure it to further split an insecure interval to potentially discover secure sub-intervals within the insecure interval.



\noindent\textbf{Optimizing iterative refinement.}
We develop two other optimizations, namely \ia and \mono, to further cut the average bisection depths. 

\textit{(1) Influence analysis.} 
When deciding which input intervals to bisect first, instead of following a random strategy, we compute the gradient or Jacobian of the output with respect to each input feature and pick the largest one as the first to bisect. The high-level intuition is that the gradient approximates the influence of the input on the output, which essentially measures the sensitivity of the output to each input feature. 

\begin{algorithm}[!hbt]
	\caption{Backward propagation for gradient interval}
    \label{alg:backward}
	\scalebox{0.75}{
		
		\begin{tabular}{|llp{9em}|}
			\hline
			\textbf{Inputs}: & \textbf{network} $\leftarrow$ tested neural network & \  \\ 
			\ & \textbf{R} $\leftarrow$ gradient mask & \ \\
			\hline
		\end{tabular}}
	\small
	\begin{spacing}{0.9}
	\begin{algorithmic}[1]
		\State // initialize upper and lower gradient bounds
		\State $g_{up} = g_{low}$ = weights[lastLayer];
		\For {layer = numlayer-1 $\mathbf{to}$ 1}
			\For {1 $\mathbf{to}$ layerSize[layer]}
				\State // $g$ is an interval containing $g_{up}$ and $g_{low}$
				\State // interval hadamard product
				\State $g$=R[layer] $\bigotimes$ $g$;
				\State // interval matrix multiplication
				\State $g$=weights[layer] $\bigodot$ $g$;
			\EndFor	
		\EndFor
		\State \Return $g$;
	\end{algorithmic}
	\end{spacing}
\end{algorithm}

Algorithm~\ref{alg:backward} shows the steps for backward computation of the input feature influence. Note that instead of working on concrete values, this version works with intervals. The basic idea is to approximate the influence caused by ReLUs. If there is no ReLU in the target DNN, the Jacobian matrix is completely determined by the weight parameters, which is independent of the input. A ReLU node's gradient can either be 0 for negative input or 1 for positive input. We use intervals to track and propagate the bounds on the gradients of the ReLU nodes during backward propagation as shown in Algorithm~\ref{alg:backward}.


We further use the estimated gradient interval to compute the smear function for an input feature~\cite{kearfott2013rigorous,kearfott1990algorithm}: $S_i(X) = max_{1\leq j \leq d}\overline{|J_{ij}|}w(X_j)$, where $J_{ij}$ denotes the gradient of input $X_j$ for output $Y_i$. For each refinement step, we bisect the $X_j$ with the highest smear value to reduce the over-approximation error as shown in Algorithm~\ref{alg:smear}.

\begin{algorithm}[!hbt]
	\caption{Using \ia to choose the most influential feature to split}
	\label{alg:smear}
    \scalebox{0.75}{
	\begin{tabular}{|ll|}
		\hline
		\textbf{Inputs}: & \textbf{network} $\leftarrow$ tested neural network  \\ 
		\ & \textbf{input} $\leftarrow$ input interval \\
		\ & \textbf{g} $\leftarrow$ gradient interval calculated by backward propagation\\
		\hline
	\end{tabular}}
	\small
	\begin{spacing}{0.9}
	\begin{algorithmic}[1]
		\For {i = 1 $\mathbf{to}$ input.length}
		\State // r is the range of each input interval
		\State $r = w(input[i])$;
		\State // e is the influence from each input to output
		\State e = $g_{up}[i]*r$;
		\If{e $>$ largest} \Comment{most effective feature}
		\State largest = e;
		\State splitFeature = i;
		\EndIf
		\EndFor
		\State \Return splitFeature;
	\end{algorithmic}
	\end{spacing}
\end{algorithm}

\textit{(2) Monotonicity.} Computing the Jacobian matrix also helps us to reason about the monotonicity property of the output for a given input interval.
In particular, for the cases where the partial derivative of $\frac{\partial F_i}{\partial X_j}$ is always positive or negative for the given input interval $X$, we can simply replace the interval $X_j$ with two concrete values $\underline{X_j}$ and $\overline{X_j}$. Because, as the DNN output is monotonic in that input interval, it is impossible for any intermediate value to cause a violation without either $\underline{X_j}$ or $\overline{X_j}$ causing one. Our empirical results in Section~\ref{sec:eval} also indicate that such monotonicity checking can help decrease the number of splits required for checking different security properties.

	\section{Implementation}
\label{sec:implementation}

\noindent\textbf{Setup.} We implement \sys in \texttt{C} and leverage \texttt{OpenBLAS}\footnote{http://www.openblas.net/} to enable efficient matrix multiplications. 
We evaluate \sys on a Linux server running Ubuntu 16.04 with 16 CPU cores and 256GB memory.

\noindent\textbf{Parallelization.} One unique advantage of \sys over other security property checking systems like Reluplex is that the interval arithmetic in the setting of verifying DNNs is highly parallelizable by nature.
During the process of \iir, newly created input ranges can be checked independently. This feature allows us to create as many threads as possible, each taking care of a specific input range, to gain significant speedup by distributing different input ranges to different workers.

However, there are two key challenges that required solving to fully leverage the benefits of parallelization. First, as shown in Section~\ref{subsec:iir}, the bisection tree is often not balanced leading to substantially different running times for different threads. We found that often several laggard threads slow down the computation, i.e., most of the available workers stay idle while only a few workers keep on refining the intervals. Second, as it is hard to predict the depth of the bisection tree for any sub-interval in advance, starting a new thread for each sub-interval may result in high scheduling overhead. To solve these two problems, we develop a dynamic thread rebalancing algorithm that can identify the potentially deeper parts of the bisection tree and efficiently redistribute those parts among other workers.
 

\noindent\textbf{Outward rounding.} The large number of floating matrix multiplications in a DNN can potentially lead to severe precision drops after rounding~\cite{goldberg1991every}. 
For example, assume that the output of one neuron is [0.00000001, 0.00000002].
If the floating-point precision is $e-7$, then it is automatically rounded up to [0.0,0.0]. 
After one layer propagation with a weight parameter of 1000, the correct output should be [0.00001, 0.00002]. However, after rounding, the output will incorrectly become [0.0, 0.0]. As the interval propagates through the neural network, more errors will accumulate and significantly affect the output precision. In fact, our tests show that some adversarial examples reported by Reluplex~\cite{katz2017reluplex} are false positives due to such rounding problem.

To avoid such issues, we adopt outward rounding in \sys. 
In particular, for every newly calculated interval or symbolic interval, we always round the bounds outward to ensure the computed output range is always a sound overestimation of the true output range. We implement outward rounding with 32-bit floats. We find that this precision is enough for verifying properties of ACAS Xu models, though it can easily be extended to 64-bit double.

	\section{Evaluation}
\label{sec:eval}

\subsection{Evaluation Setup}
In the evaluation, we consider two general categories of DNNs, deployed for handling two different tasks.

The first category is airborne collision avoidance system (ACAS) crucial for alerting and preventing the collisions between aircraft. We focus our evaluation on ACAS Xu models for collision avoidance in unmanned aircraft~\cite{kochenderfer2012next}.

The second category includes the models deployed to recognize hand-written digit from the MNIST dataset. Our preliminary results demonstrate that \sys can also scale to larger networks that the solver-based verification tools often struggle to check.

\begin{figure}[!htb]
\centering
	\includegraphics[width=0.7\columnwidth]{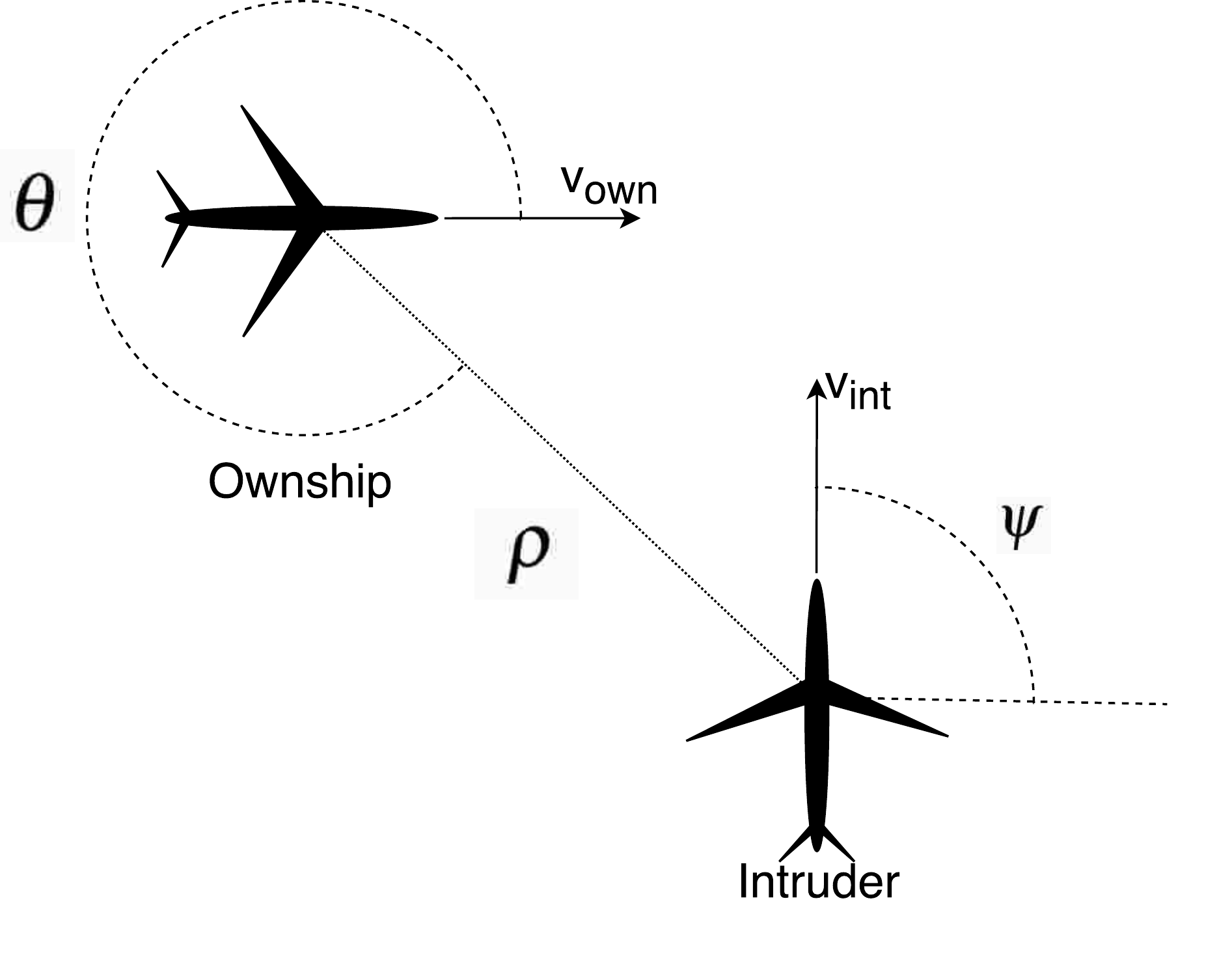}
	\caption{\small Horizontal view of ACAS Xu operating scenarios.}
	\label{acas_inputs}
\end{figure}

\noindent\textbf{ACAS Xu.} 
The ACAS Xu system consists of forty-five different NN models. 
Each network is composed of an input layer taking five inputs, an output layer generating five outputs, and six hidden layers with each containing fifty neurons. 
As shown in Figure~\ref{acas_inputs}, five inputs include $\{\rho, \theta,\psi,v_{own}, v_{int}\}$.
In particular, $\rho$ denotes the distance between ownship and intruder, $\theta$ denotes the heading direction angle of ownship relative to the intruder, $\psi$ denotes the heading direction angle of the intruder relative to ownship, $v_{own}$ is the speed of ownship, and $v_{int}$ is the speed of intruder.
Output of the NN includes \{\textit{COC}, \textit{weak left}, \textit{weak right}, \textit{strong left}, \textit{strong right}\}. 
\textit{COC} denotes clear of conflict, \textit{weak left} means heading left with angle $1.5^{\omicron}$/s, \textit{weak right} means heading right with angle $1.5^{\omicron}$/s, \textit{strong left} is heading left with angle $3.0^{\omicron}$/s, and \textit{strong right} denotes heading right with angle $3.0^{\omicron}$/s. 
Each output in NN corresponds to the score for this action (minimal for the best).

\noindent\textbf{MNIST.} For classifying hand-written digits, we test a neural network with 784 inputs, 10 outputs, and two hidden layers. Each intermediate layer has 512 neurons. On the MNIST test dataset, it can achieve 98.28\% accuracy for classification.

\begin{table*}[!t]
\setlength{\tabcolsep}{8pt}
\centering
	\begin{tabular}[width = \textwidth]{|c |c |c |l |l | c|}
		\hline
Source & Properties & Networks & Reluplex Time (sec) & \sys Time (sec) & Speedup\\ 
\hline
\multirow{10}{*}{\makecell{Security \\Properties \\ from \cite{katz2017reluplex} }} 
& $\phi_1$ & 45 & $>$443,560.73* & 14,603.27 & $>$30$\times$\\

& $\phi_2$ & 34$^*$\footnotemark & 123,420.40 &  117,243.26 & 1$\times$ \\

& $\phi_3$ & 42 & 35,040.28 & 19,018.90 & 2$\times$\\

& $\phi_4$ & 42 & 13,919.51 & 441.97 & 32$\times$\\

& $\phi_5$ & 1 & 23,212.52 & 216.88 & 107$\times$\\

& $\phi_6$ & 1 & 220,330.82 & 46.59 & 4729$\times$\\

& $\phi_7$ & 1 & $>$86400.0* & 9,240.29 & $>$9$\times$\\

& $\phi_8$ & 1 & 43,200.01 & 40.41 & 1069$\times$\\

& $\phi_9$ & 1 & 116,441.97 & 15,639.52 & 7$\times$\\

& $\phi_{10}$ & 1 & 23,683.07 & 10.94 & 2165$\times$\\
\hline\hline
\multirow{5}{*}{\makecell{Additional \\Security \\Properties}} 
& $\phi_{11}$ & 1 & 4,394.91 & 27.89 & 158$\times$\\

& $\phi_{12}$ & 1 & 2,556.28 & 0.104 & 24580$\times$\\

& $\phi_{13}$ & 1 & $>$172,800.0* & 148.21 & $>$1166$\times$\\

& $\phi_{14}$ & 2 & $>$172,810.86* & 288.98 & $>$598$\times$\\

& $\phi_{15}$ & 2 & 31,328.26 & 876.80 & 36$\times$\\
\hline
        \multicolumn{6}{l}{\footnotesize * Reluplex uses different timeout thresholds for different properties. }
	\end{tabular}
	\caption{\sys's performance at verifying properties of ACAS Xu compared with Reluplex. $\phi_1$ to $\phi_{10}$ are the properties proposed in Reluplex~\cite{katz2017reluplex}. $\phi_{11}$ to $\phi_{15}$ are our additional properties.}
	\label{properties}
\end{table*}

\subsection{Performance on ACAS Xu Models}
In this section, we first present a detailed comparison of \sys and Reluplex in terms of the verification performance. 
Then, we compare \sys with a state-of-the-art adversarial attack on DNNs, Carlini-Wagner~\cite{carlini2017towards}, showing that on average \sys can consistently find 50\% more adversarial examples. 
Finally, we show that \sys can accurately narrow down all possible adversarial ranges and therefore provide more insights on the distribution of adversarial corner-cases.

\noindent\textbf{Comparison to Reluplex.}
Table~\ref{properties} compares the time taken by \sys with that of Reluplex for verifying ten original properties described in their paper~\cite{katz2017reluplex}.
In addition, we include the experimental results for five new security properties. The detailed description of each property is in the Appendix. 
Table~\ref{properties} shows that \sys always outperforms Reluplex at checking all fifteen security properties.
For the properties on which Reluplex times out, \sys is able to terminate in significantly shorter time.
On average, \sys achieves up to $200\times$ speedup over Reluplex. 

\begin{table}[!t]
\setlength{\tabcolsep}{6pt}
\footnotesize
\centering
	\begin{tabular}{|c|c|c|c|c|}
		\hline
		\textbf{\# Seeds} & \textbf{CW} & \textbf{CW Miss} & \textbf{\sys{}} & \textbf{\sys Miss} \\ 
		\hline
		50 & 24/40 & 40.0\% & 40/40 & 0\% \\
		\hline
		40 & 21/40 & 47.5\% & 40/40 & 0\% \\
		\hline
		30 & 17/40 & 58.5\% & 40/40 & 0\% \\
		\hline
		20 & 10/40 & 75.0\% & 40/40 & 0\% \\
		\hline
		10 & 6/40 & 85.0\% & 40/40 & 0\% \\
		\hline
	\end{tabular}
	\small
	\caption{The number of adversarial inputs CW can find compared to \sys on 40 adversarial ACAS Xu properties. The third column shows the percentage of adversarial properties CW failed to find.}
	\label{CW1}
\end{table}

\noindent\textbf{Finding adversarial inputs.}
In terms of the number of adversarial examples detected, \sys also outperforms the popular attacks using gradients to find adversarial examples.
Here, we compare \sys to the Carlini and Wagner (CW) attack~\cite{carlini2017towards}, a state-of-the-art gradient-based attack that minimizes specialized CW loss function. 

As gradient-based attacks start from a seed input and iteratively looking for adversarial examples, the choice of seeds may highly influence the success of the attack at finding adversarial inputs. Therefore, we try different randomly picked seed inputs to facilitate the input generation process. 
Note that our technique in \sys does not need any seed input. Thus it is not restricted by the potentially undesired starting seed and can fully explore the input space.
As shown in Table \ref{CW1}, on average, CW misses 61.2\% number of models, which do have adversarial inputs exist that CW fails to find.

\noindent\textbf{Narrowing down adversarial ranges.}
A unique feature of \sys is that it can isolate adversarial ranges of inputs from the non-adversarial ones. 
This is useful because it allows a DNN designer to potentially isolate and avoid adversarial ranges with a given precision (\eg $e-6$ or smaller). 
Here we set the precision to be $e-6$, i.e., we allow splitting of the intervals into smaller sub-intervals unless their length becomes less than $e-6$. 
Table~\ref{tab:narrow} shows the results of the three different properties that we checked. 
For example, property $S_1$ specifies \texttt{model\_4\_1} should output strong right with input range $\rho = [400,10000]$, $\theta = 0.2$, $\psi = -3.09$, $v_{own} = 10$, and $v_{int} = 10$. 
For this property, \sys splits the input ranges into 262,144 smaller sub-intervals and is able to prove that 163,915 sub-intervals are safe.
\sys also finds that $\rho = [400, 6402.36]$ does not contain any adversarial inputs while $\rho = [6402.36,10000]$ is adversarial.

\begin{table}[!t]
\setlength{\tabcolsep}{4pt}
\footnotesize
\centering
	\begin{tabular}{|c|c|c|c|c|}
		\hline
		\textbf{P} & \textbf{Adv Range} & \textbf{Adv} & \textbf{Timeout} & \textbf{Non-adv}\\ 
		\hline
		$S_1$ & $[6402.36,10000]$ & 98229 & 1 & 163915\\
		\hline
		$S_2$ & $[-0.2,-0.186]$ and $[-0.103,0]$ & 18121 & 2 & 14645\\
		\hline
		$S_3$ & $[-0.1,0.0085]$ & 17738 & 1 & 15029\\
		\hline
	\end{tabular}
	\small
	\caption{The second column shows the input ranges containing at least one adversarial input, while the rest of ranges are found by \sys to be non-adversarial. The last three columns show the number of total sub-intervals checked by \sys with a precision of $e-6$.}
	\label{tab:narrow}
\end{table}

\subsection{Preliminary Tests on MNIST Model}
\begin{figure}[!hbt]
\centering
	\includegraphics[width=0.8\columnwidth]{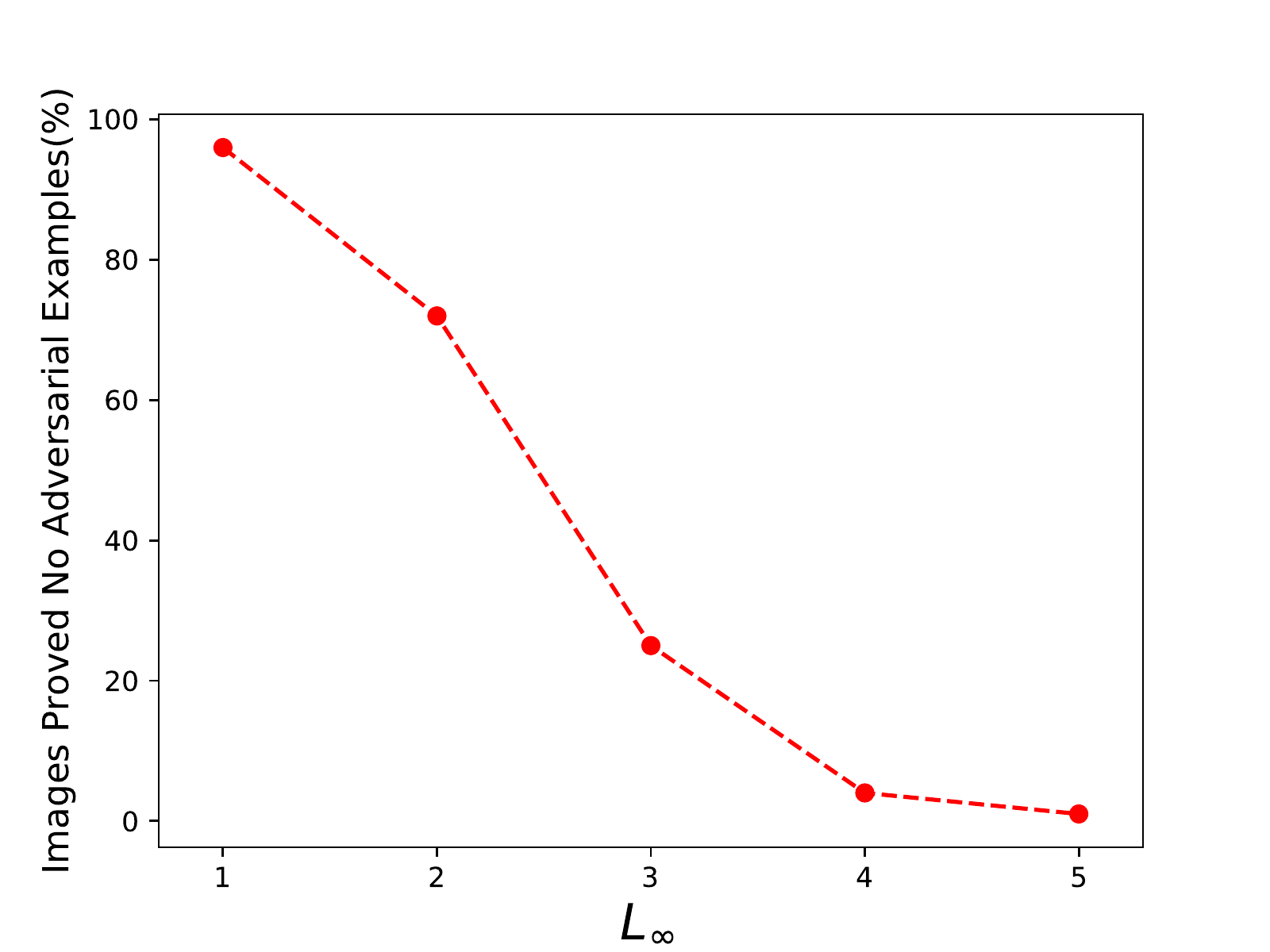}
	\caption{Percentage of images proved to be non-adversarial with $L_{\infty}=1,2,3,4,5$ by \sys on MNIST test model out of 1000 random test MNIST images.}
	\label{fig:MNIST}
\end{figure}

Besides ACAS Xu, we also test \sys on an MNIST model that achieves decent accuracy (98.28\%).
Given a particular seed image, we allow arbitrary perturbations to every pixel value while bounding the total perturbation by the $L_\infty$ norm.
In particular, \sys can prove 956 seed images to be safe for $|X|_{\infty}\leq1$ and 721 images safe for $|X|_{\infty}\leq2$ respectively out of 1000 randomly selected test images. Figure~\ref{fig:MNIST} shows the detailed results. As the norm is increased, the percentage of images that have no adversarial perturbations drops quickly to 0.
Note that we get more timeouts as the $L_\infty$ norm increase. We believe that we can further optimize our system to work on GPUs to minimize such timeouts and verify properties with larger norm bounds. 

\footnotetext{We remove model\_4\_2 and model\_5\_3 because Reluplex found incorrect adversarial examples due to rounding problems (these models do not have any adversarial case).}

\subsection{Optimizations}

In this subsection, we evaluate the effectiveness of the optimizations proposed in Section~\ref{sec:methodology} compared to the naive interval extension with \iir. The results are shown in Table~\ref{savings}.

\begin{table}[!hbt]
\setlength{\tabcolsep}{6pt}
\footnotesize
\centering
	\begin{tabular}[width = 0.48\textwidth]{|c|c|c|c|}
		\hline
		\textbf{Methods} & \textbf{Deepest Dep (\%)} & \textbf{Avg Dep (\%)} & \textbf{Time (\%)} \\ 
		\hline
		S.C.P & 42.06 & 49.28 & 99.99 \\
		\hline
		I.A. & 10.65 & 10.85 & 96.04 \\
		\hline
		Mono & 0.325 & 0.497 & 16.91 \\
		\hline
	\end{tabular}
	\small
	\caption{The percentages of the deepest depth, average depth, and average running time improvement caused by the three main components of \sys: \scp, \ia, and \mono compared to the naive interval analysis.}
	\label{savings}
\end{table}

\noindent\textbf{Symbolic interval propagation.} Table~\ref{savings} shows that \scp saves the deepest and average depth of bisection tree (Figure~\ref{fig:bisectionTree}) by up to 42.06\% and 49.28\%, respectively, over naive interval extension. 


\noindent\textbf{Influence analysis.} As one of the optimizations used in iterative refinement, influence analysis helps prioritize splitting of the most influential input to the output. Compared to the sequential splitting features, influence-analysis-based splitting reduces the average depth by 10.85\% and thus cut down the running time by up to 96.04\%.

\noindent\textbf{Monotonocity.} The improvements from using \mono are relatively smaller in terms of tree depth.
However, it can still reduce the average running time by 16.91\% on average, especially when the average depth is high.

	\section{Related Work}
\label{sec:related}

\noindent\textbf{Adversarial machine learning.} Several recent works have shown that even the state-of-the-art DNNs can be easily fooled by adding small carefully crafted human-imperceptible perturbations to the original inputs~\cite{szegedy2013intriguing, goodfellow2014explaining, moosavi2016deepfool, carlini2017towards}. This has resulted in an arms race among researchers competing to build more robust networks and design more efficient attacks~\cite{goodfellow2014explaining, papernot2016practical, kurakin2016adversarial, carlini2017towards, liu2016delving, nguyen2015deep, xu2016automatically}. However, most of the defenses are restricted to only one type of adversaries/security properties (\eg overall perturbations bounded by some norms) even though other researchers have shown that other semantics-preserving changes like lightning changes, small occlusions, rotations, etc. can also easily fool the DNNs~\cite{pei2017deepxplore, tian2017deeptest, pei2017towards, engstrom2017rotation}.  However, none of these attacks can provide any provable guarantees about the non-existence of adversarial examples for a given neural network. Unlike these attacks, \sys can provide a provable security analysis of given input ranges, systematically narrowing down and detecting all adversarial ranges.

\noindent\textbf{Verification of machine learning systems.} 
Recently, several projects~\cite{katz2017reluplex, ehlers2017formal, huang2017safety} have used customized SMT solvers for verifying security properties of DNNs. However, such techniques are mostly limited by the scalability of the solver. Therefore, they tend to incur significant overhead~\cite{katz2017reluplex} or only provide weaker guarantees~\cite{huang2017safety}. By contrast, \sys uses interval-based techniques and significantly outperforms the state-of-the-art solver-based systems like Reluplex~\cite{katz2017reluplex}.

Kolter et al.~\cite{kolter2017provable} and Raghunathan et al.~\cite{raghunathan2018certified} transform the verification problem into a convex optimization problem using relaxations to over-approximate the outputs of ReLU nodes. Similarly, Gehr et al.~\cite{gehr2018ai} leverages zonotopes for approximating each ReLU outputs. Dvijotham et al.~\cite{dvijotham2018dual} transform the verification problem into an unconstrained dual formulation using Lagrange relaxation and use gradient-descent to solve the optimization problem. However, all of these works focus on simply over-approximating the total number of potential adversarial violations without trying to find concrete counterexamples. Therefore, they tend to suffer from high false positive rates unless the underlying DNN's training algorithm is modified to minimize such violations. By contrast, \sys can find concrete counterexamples as well as verify security properties of pre-trained DNNs. 

Recently, Mixed Integer Linear programming (MILP) solvers combined with gradient descent have also been proposed for verification of DNNs~\cite{dutta2018output,dutta2018learning}. Integrating our interval analysis together with such approaches is an interesting future research problem. 

Verivis\cite{pei2017towards} by Pei et al. is a black-box DNN verification system that leverages the discreteness of image pixels. However, unlike ReluVal, it cannot verify non-existence of norm-based adversarial examples.

\noindent\textbf{Interval optimization.} 
Interval analysis has shown great success in many application domains including non-linear equation solving and global optimization problems~\cite{jaulin1993guaranteed,moore1962interval, moore2003interval}. Due to its ability to provide rigorous bounds on the solutions of an equation, many numerical optimization problems~\cite{ishii2015scalable,vaidyanathan1994global} leveraged interval analysis to achieve a near-precise approximation of the solutions. 
We note that the computation inside NN is mostly a sequence of simple linear transformations with nonlinear activation functions.
These computations thus highly resemble those in traditional domains where interval analysis has been shown to be successful.
Therefore, based on the foundation of interval analysis laid by Moore et al.~\cite{intervalbook,moore1979methods}, we leverage interval analysis for analyzing the security properties of DNNs.

	\section{Future Work and Discussion}
\label{discussion}

\noindent{\textbf{Supporting other activation functions.} Interval extension can, in theory, be applied to any activation function that maintains inclusion isotonicity and Lipschitz continuity. As mentioned in Section 4, most popular activation functions (e.g., tanh, sigmoid) satisfy these properties. To support these activation functions, we need to adapt the symbolic interval propagation process. We plan to explore this as part of future work. Our current prototype implementation of symbolic interval propagation supports several common piece-wise linear activation functions (e.g., regular ReLU, Leaky ReLU, and PReLU).

\noindent{\textbf{Supporting other norms besides $L_{\infty}$.} While interval arithmetic is most immediately applicable to $L_{\infty}$, other norms (e.g., $L_{2}$ and $L_{1}$) can also be approximated using intervals. Essentially, $L_{\infty}$ allows the most flexible perturbations and the perturbations bounded by other norms like $L_{2}$ are all subsets of those allowed by the corresponding $L_{\infty}$  bound. Therefore, if \sys can verify the absence of adversarial examples for a DNN within an infinite norm bound, the DNN is also guaranteed to be safe for the corresponding p-norm (p=1/2/3..) bound. If \sys identifies adversarial subintervals for an infinite norm bound, we can iteratively check whether any such subinterval lies within the corresponding p-norm bound. If not, we can declare the model to contain no adversarial examples for the given p-norm bound. We plan to explore this direction in future.

\noindent{\textbf{Improving DNN Robustness.} The counterexamples found by \sys can be used to increase the robustness of a DNN through adversarial training. Specific, we can add the adversarial examples detected by ReluVal to the training dataset and retrain the model. Also, a DNN's training process can further be changed to incorporate ReluVal's interval analysis for improved robustness. Instead of training on individual samples, we can convert the training samples into intervals and change the training process to minimize losses for these intervals instead of individual samples. We plan to pursue this direction as future work.

\section{Conclusion}
\label{sec:conclusion}

Although this paper focuses on verifying security properties of DNNs, \sys itself is a generic framework that can efficiently leverage interval analysis to understand and analyze the DNN computation. In the future, we hope to develop a full-fledged DNN security analysis tool based on \sys, just like traditional program analysis tools, that can not only efficiently check arbitrary security properties of DNNs but can also provide insights into the behaviors of hidden neurons with rigorous guarantees.

In this paper, we designed, developed, and evaluated \sys, a formal security analysis system for neural networks. We introduced several novel techniques including symbolic interval arithmetic to perform formal analysis without resorting to SMT solvers. \sys performed 200 times faster on average than the current state-of-art solver-based approaches. 
	
	\section{Acknowledgements}
\label{acknowledgements}

We thank Chandrika Bhardwaj, Andrew Aday, and the anonymous reviewers for their constructive and valuable feedback. This work is sponsored in part by NSF grants CNS-16-17670, CNS-15-63843, and CNS-15-64055; ONR grants N00014-17-1-2010, N00014-16-1-2263, and N00014-17-1-2788; and a Google Faculty Fellowship. Any opinions, findings, conclusions, or recommendations expressed herein are those of the authors, and do not necessarily reflect those of the US Government, ONR, or NSF.

	\small
	\bibliographystyle{abbrv}
	\bibliography{ref}
	
	\appendix
\section{Appendix: Formal Definitions for ACAS Xu Properties $\phi_1$ to $\phi_{15}$}
\textbf{\textit{Inputs.}} Inputs for each ACAS Xu DNN model are:

$\rho$: the distance between ownship and intruder;

$\theta$: the heading direction angle of ownship relative to intruder;

$\psi$: heading direction angle of intruder relative to ownship;

$v_{own}$: speed of ownshipe;

$v_{int}$: speed of intruder;

\noindent \textbf{\textit{Outputs.}} Outputs for each ACAS Xu DNN model are:

\textit{COC}: Clear of Conflict;

\textit{weak left}: heading left with angle $1.5^{\omicron}$/s;

\textit{weak right}: heading right with angle $1.5^{\omicron}$/s;

\textit{strong left}: heading left with angle $3.0^{\omicron}$/s;

\textit{strong right}: heading right with angle $3.0^{\omicron}$/s.

\noindent \textbf{\textit{45 Models.}} There are 45 different models indexed by two extra inputs $a_{prev}$ and $\tau$, model\_x\_y means the model used when $a_{prev}=x$ and $\tau=y$ :

$a_{prev}$: previous action indexed as \{\textit{COC}, \textit{weak left}, \textit{weak right}, \textit{strong left}, \textit{strong right}\}.

$\tau$: time until loss of vertical separation indexed as \{0, 1, 5, 10, 20, 40, 60, 80, 100\}

\noindent \textbf{\textit{Property $\phi_1$:}} If the intruder is distant and is significantly slower than the ownship, the score of a COC advisory will always be below a certain fixed
threshold.

Tested on: all 45 networks.

Input ranges: $\rho \geq 55947.691$, $v_{own} \geq 1145$, $v_{int} \leq 60$.

Desired output: the output of COC is at most 1500.

\noindent \textbf{\textit{Property $\phi_2$:}} If the intruder is distant and is significantly slower than the ownship, the score of a COC advisory will never be maximal.

Tested on: model\_x\_y, $x\geq2$, except model\_5\_3 and model\_4\_2

Input ranges: $\rho \geq 55947.691$, $v_{own} \geq 1145$, $v_{int} \leq 60$.

Desired output: the score for COC is not the maximal score.

\noindent \textbf{\textit{Property $\phi_3$:}}  If the intruder is directly ahead and is moving towards the ownship, the score for COC will not be minimal.

Tested on: all models except model\_1\_7, model\_1\_8 and model\_1\_9

Input ranges: $1500 \leq \rho \leq 1800$, $-0.06\leq \theta \leq 0.06$, $\psi \geq 3.10$, $v_{own} \geq 980$, $v_{int} \geq 960$.

Desired output: the score for COC is not the minimal score.

\noindent \textbf{\textit{Property $\phi_4$:}}  If the intruder is directly ahead and is moving away from the ownship but at a lower speed than that of the ownship, the score for COC will not be minimal.

Tested on: all models except model\_1\_7, model\_1\_8 and model\_1\_9

Input ranges: $1500 \leq \rho \leq 1800$, $-0.06\leq \theta \leq 0.06$, $\psi = 0$, $v_{own} \geq 1000$, $700\leq v_{int} \leq 800$.

Desired output: the score for COC is not the minimal score.

\noindent \textbf{\textit{Property $\phi_5$:}} If the intruder is near and approaching from the left, the network advises “strong right”.

Tested on: model\_1\_1

Input ranges: $250 \leq \rho \leq 400$, $0.2\leq \theta \leq 0.4$, $-3.141592 \leq \psi \leq -3.141592+0.005$, $100 \leq v_{own} \leq 400$, $0\leq v_{int} \leq 400$.

Desired output: the score for ``strong right'' is the minimal score.

\noindent \textbf{\textit{Property $\phi_6$:}} If the intruder is sufficiently far away, the network advises COC.

Tested on: model\_1\_1

Input ranges: $12000 \leq \rho \leq 62000$, $(0.7\leq \theta \leq 3.141592) \cup (-3.141592\leq \theta \leq -0.7)$, $-3.141592 \leq \psi \leq -3.141592+0.005$, $100 \leq v_{own} \leq 1200$, $0\leq v_{int} \leq 1200$.

Desired output: the score for COC is the minimal score.

\noindent \textbf{\textit{Property $\phi_7$:}} If vertical separation is large, the network will never advise a strong turn

Tested on: model\_1\_9

Input ranges: $0 \leq \rho \leq 60760$, $-3.141592\leq \theta \leq 3.141592$, $-3.141592 \leq \psi \leq 3.141592$, $100 \leq v_{own} \leq 1200$, $0\leq v_{int} \leq 1200$.

Desired output: the scores for ``strong right'' and ``strong left'' are never the minimal scores.

\noindent \textbf{\textit{Property $\phi_8$:}} For a large vertical separation and a previous 	``weak left'' advisory, the network will either output COC or continue advising ``weak left.''

Tested on: model\_2\_9

Input ranges: $0 \leq \rho \leq 60760$, $-3.141592\leq \theta \leq -0.75 \cdot 3.141592$, $-0.1 \leq \psi \leq 0.1$, $600 \leq v_{own} \leq 1200$, $600\leq v_{int} \leq 1200$.

Desired output: the score for ``weak left'' is minimal or the score for COC is minimal.

\noindent \textbf{\textit{Property $\phi_9$:}} Even if the previous advisory was ``weak right,'' the presence of a nearby intruder will cause the network to output a ``strong left'' advisory
instead.

Tested on: model\_3\_3

Input ranges: $2000 \leq \rho \leq 7000$, $0.7\leq \theta \leq 3.141592$, $-3.141592 \leq \psi \leq -3.141592+0.01$, $100 \leq v_{own} \leq 150$, $0\leq v_{int} \leq 150$.

Desired output: the score for ``strong left'' is minimal.

\noindent \textbf{\textit{Property $\phi_{10}$:}} For a far away intruder, the network advises COC.

Tested on: model\_4\_5

Input ranges: $36000 \leq \rho \leq 60760$, $0.7\leq \theta \leq 3.141592$, $-3.141592 \leq \psi \leq -3.141592+0.01$, $900 \leq v_{own} \leq 1200$, $600\leq v_{int} \leq 1200$.

Desired output: the score for COC is minimal.

\noindent \textbf{\textit{Property $\phi_{11}$:}} If the intruder is near and approaching from the left but the vertical separation is comparably large, the network still tend to advise ``strong right'' more than COC.

Tested on: model\_1\_1

Input ranges: $250 \leq \rho \leq 400$, $0.2\leq \theta \leq 0.4$, $-3.141592 \leq \psi \leq -3.141592+0.005$, $100 \leq v_{own} \leq 400$, $0\leq v_{int} \leq 400$.

Desired output: the score for ``strong right'' is always smaller than COC.

\noindent \textbf{\textit{Property $\phi_{12}$:}} If the intruder is distant and is significantly slower than the ownship, the score of a COC advisory will be the minimal.

Tested on: model\_3\_3

Input ranges: $\rho \geq 55947.691$, $v_{own} \geq 1145$, $v_{int} \leq 60$.

Desired output: the score for COC is the minimal score.

\noindent \textbf{\textit{Property $\phi_{13}$:}} For a far away intruder but the vertical distance are small, the network always advises COC no matter the directions are.

Tested on: model\_1\_1

Input ranges: $60000 \leq \rho \leq 60760$, $-3.141592\leq \theta \leq 3.141592$, $-3.141592 \leq \psi \leq 3.141592$, $0 \leq v_{own} \leq 360$, $0\leq v_{int} \leq 360$.

Desired output: the score for COC is the minimal.

\noindent \textbf{\textit{Property $\phi_{14}$:}} If the intruder is near and approaching from the left and vertical distance is small, the network always advises strong right no matter previous action is strong right or strong left.

Tested on: model\_4\_1, model\_5\_1

Input ranges: $250 \leq \rho \leq 400$, $0.2\leq \theta \leq 0.4$, $-3.141592 \leq \psi \leq -3.141592+0.005$, $100 \leq v_{own} \leq 400$, $0\leq v_{int} \leq 400$.

Desired output: the score for ``strong right'' is always the minimal.

\noindent \textbf{\textit{Property $\phi_{15}$:}} If the intruder is near and approaching from the right and vertical distance is small, the network always advises strong left no matter previous action is strong right or strong left.

Tested on: model\_4\_1, model\_5\_1

Input ranges: $250 \leq \rho \leq 400$, $-0.4\leq \theta \leq -0.2$, $-3.141592 \leq \psi \leq -3.141592+0.005$, $100 \leq v_{own} \leq 400$, $0\leq v_{int} \leq 400$.

Desired output: the score for ``strong left'' is always the minimal.

\noindent \textbf{\textit{Property $S_1$:}}

Tested on: model\_4\_1

Input ranges: $400 \leq \rho \leq 10000$, $\theta = 0.2$, $\psi = -3.141592+0.005$, $v_{own} = 10$, $v_{int} = 10$.

Desired output: the score for ``strong right'' is the minimal.

\noindent \textbf{\textit{Property $S_2$:}}

Tested on: model\_4\_1

Input ranges: $\rho = 400$, $-0.2 \leq \theta \leq 0$, $\psi = -3.141592+0.005$, $v_{own} = 1000$, $v_{int} = 1000$.

Desired output: the score for ``strong right'' is the minimal.

\noindent \textbf{\textit{Property $S_3$:}}

Tested on: model\_1\_2

Input ranges: $\rho = 400$, $-0.1 \leq \theta \leq 0.1$, $\psi = -3.141592$, $v_{own} = 500$, $v_{int} = 600$.

Desired output: the score for ``strong right'' is the minimal.

\end{document}